# Projection based Active Gaussian Process Regression for Pareto Front Modeling

Zhengqi Gao[1], Jun Tao[1], Yangfeng Su[2], Dian Zhou[3], and Xuan Zeng[1]
[1]ASIC & System State Key Lab, School of Microelectronics, Fudan University, Shanghai, China
[2]School of Mathematical Sciences, Fudan University, Shanghai, China
[3]Department of EE, University of Texas at Dallas, Dallas, USA

*Abstract*—Pareto Front (PF) modeling is essential in decision making problems across all domains such as economics, medicine or engineering. In Operation Research literature, this task has been addressed based on multi-objective optimization algorithms. However, without learning models for PF, these methods cannot examine whether a new provided point locates on PF or not. In this paper, we reconsider the task from Data Mining perspective. A novel projection based active Gaussian process regression (P-aGPR) method is proposed for efficient PF modeling. First, P-aGPR chooses a series of projection spaces with dimensionalities ranking from low to high. Next, in each projection space, a Gaussian process regression (GPR) model is trained to represent the constraint that PF should satisfy in that space. Moreover, in order to improve modeling efficacy and stability, an active learning framework has been developed by exploiting the uncertainty information obtained in the GPR models. Different from all existing methods, our proposed P-aGPR method can not only provide a generative PF model, but also fast examine whether a provided point locates on PF or not. The numerical results demonstrate that compared to state-of-the-art passive learning methods the proposed P-aGPR method can achieve higher modeling accuracy and stability.

*Keywords—Projection, Gaussian process regression, active learning, Pareto Front modeling*

## I. INTRODUCTION

Many real-world decision making problems in biology [1]-[2], economics [3], medicine [4] or engineering [5]-[8], etc., involve optimizing several performance metrics simultaneously with respect to the design variables. For instance, in economics [3], we aim to allocate the scarce capital among the investment opportunities so that we can achieve high profit and low risk at the same time. However, in reality we cannot optimize all the metrics simultaneously, and trade-offs among metrics always exist [1]-[8].

The aforementioned metrics' trade-offs can be well described by *Pareto Front*. Accurate Pareto Front modeling is of crucial importance for optimal decision making. For instance, in order to optimize a large-scale analog circuit system, we have to extract the Pareto Front of each individual circuit block. Next, the system-level analog circuit optimization can be carried out with these Pareto Front models working as constraints [8].

Conventionally in Operation Research literature, Pareto Front modeling is based on multi-objective optimization [9]-[13]. Most existing multi-objective optimization algorithms can be classified into two categories: (1) evolutionary methods [9]-[10], and (2) deterministic (non-evolutionary) methods [11]-[13]. Evolutionary methods, such as MOEA/D [9] or NSGA-II [10], start from several randomly generated *members*, and in each iteration new *offspring* will be generated based on *reproduction* and *mutation* operators. Next, those offspring with high *fitness* value will be selected out as *parents* for the next generation. After several generations, the Pareto Front can be well approximated by the generated offspring. On the other hand, deterministic methods (i.e., the second category), such as NBI [11]-[12] or SPO [13], turn the multiple objective functions into a single one by introducing a weight vector. It has been shown that solving the single-objective optimization will yield one *Pareto optimal* solution for the original problem [11]-[13]. Therefore, the Pareto Front can be obtained by repeatedly altering the weight vector and solving the corresponding scalarized optimization problem. All the aforementioned multi-objective optimization methods finally yield a data set made up of several points located on Pareto Front. However, if a new point is provided, we still cannot fast examine whether it locates on Pareto Front or not because we do not have a model describing the Pareto Front.

To address the aforementioned issue (i.e., fast examining whether a new provided point locates on PF or not) and accurately model the Pareto Front, in this paper, we propose a novel projection based active Gaussian process regression (P-aGPR) method. The proposed P-aGPR method is motivated by the fact that if the projection of a provided point onto some low dimensional space is not contained in the same projection of Pareto Front, this point cannot locate on the true Pareto Front in the original space. Based on this observation, we build several Gaussian process regression [14] (GPR) models in a series of chosen projection spaces with dimensionalities ranking from low to high. Namely, each GPR model represents a constraint that the Pareto Front should satisfy in the corresponding projection space. Next, we can iterate through the GPR models from low dimensionality to high and examine whether the provided point satisfies the constraints posed by the GPR models. When violation occurs, we can assert that the provided point doesn't locate on Pareto Front and stop immediately.

In order to further improve modeling accuracy and stability, we develop an *active learning* [15]-[19] framework to fully exploit the uncertainty information obtained in GPR models. To be more specific, the aforementioned GPR models are first initialized with a few samples. Next, we find the positions where GPR models have high predictive uncertainty, and consequently a modified NBI [11]-[12] method is proposed to query new PF samples around these positions. Then the GPR models are re-calibrated with these new obtained samples. By repeatedly querying new PF samples and retraining GPR models until convergence, we can generatively approximate the whole Pareto Front. As will be demonstrated in the numerical examples in Section V, the proposed P-aGPR method can model the Pareto Front with fewer samples and higher stability compared with state-of-the-art *passive learning* methods.

The remainder of this paper is organized as follows. In Section II, we give our problem formulation and summarize the relations of Pareto Front modeling with other existing problems. Next, we develop the proposed P-aGPR method in Section III. We further discuss several important implementation issues in Section IV. The efficacy of our proposed P-aGPR method is demonstrated by four testbench functions in Section V. Finally, we conclude in Section VI.

## II. Preliminaries

In sub-section II.A, we first introduce the notations, and give our problem formulation. Next, in sub-section II.B, we summarize the relations of Pareto Front modeling with other existing problems in Data Mining.

### A. Problem Formulation

We use $\mathbf{x} = [x_1\ x_2\ \cdots\ x_d]^T \in \Omega \subseteq \Re^d$ to represent the design variables, where $\Omega$ represents the reasonable design space in $\Re^d$. Similarly, $\mathbf{f}(\mathbf{x}) = [f_1(\mathbf{x})\ f_2(\mathbf{x}) \cdots f_m(\mathbf{x})]^T \in \Theta_m \subseteq \Re^m$ is used to denote the metrics of interests, where $\Theta_m$ is the feasible region in metric space. Without loss of generality, in this paper, we assume that a good design should achieve small values in all $m$ metrics. If some metric doesn't satisfy this requirement (e.g., a good capital investment in economics should have large profit), we can define its negative instead (e.g., we define negative profit as a metric). In reality, intrinsic trade-offs exist among different metrics [1]-[8], and we cannot optimize all metrics $\mathbf{f}(\mathbf{x})$ simultaneously. These metrics' trade-offs can be well characterized by Pareto Front.

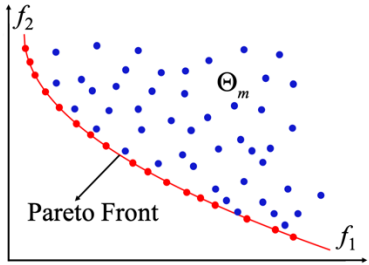

Fig. 1. Illustration of a 2-dimensional Pareto Front (i.e., $m = 2$), $f_1$ and $f_2$ deontes two metrics of interests. Both red and blue dots are metric vectors in the feasible metrics space $\Theta_m$, and the red dots locate on Pareto Front.

We first define a metric vector $\mathbf{f}(\mathbf{x}^{(a)})$ *dominates* another metric vector $\mathbf{f}(\mathbf{x}^{(b)})$ as:

$$\mathbf{f}\left(\mathbf{x}^{(a)}\right) \neq \mathbf{f}\left(\mathbf{x}^{(b)}\right) \quad \text{and} \quad \mathbf{f}\left(\mathbf{x}^{(a)}\right) \leq \mathbf{f}\left(\mathbf{x}^{(b)}\right), \tag{1}$$

where $\mathbf{f}(\mathbf{x}^{(a)}) \leq \mathbf{f}(\mathbf{x}^{(b)})$ denotes the elementwise inequalities $f_i(\mathbf{x}^{(a)}) \leq f_i(\mathbf{x}^{(b)})$, $i = 1, 2, \cdots, m$. If there is no such $\mathbf{x} \in \Omega$ that $\mathbf{f}(\mathbf{x})$ can dominate $\mathbf{f}(\mathbf{x}^*)$, design $\mathbf{x}^*$ is called *Pareto optimal*, and *Pareto Set* (PS) is defined as the set made up of all Pareto optimal designs, i.e.,

$$\text{PS} = \left\{\mathbf{x}^* \in \Omega \middle| \forall \mathbf{x} \in \Omega,\ \mathbf{f}(\mathbf{x}) \text{ does not dominate } \mathbf{f}\left(\mathbf{x}^*\right)\right\}. \tag{2}$$

*Pareto Front* (PF) is defined as the corresponding metrics set of the Pareto optimal designs. Moreover, in this paper, we assume that we only care about the PF with metrics upper bounded by given metrics' specifications, i.e.,

$$\text{PF} = \left\{\mathbf{f}\left(\mathbf{x}^*\right) \in \Theta_m \middle| \mathbf{x}^* \in \text{PS},\ \mathbf{f}\left(\mathbf{x}^*\right) \leq \mathbf{f}_{\max}\right\}, \tag{3}$$

where $\mathbf{f}_{\max} = [f_{1,\max}\ f_{2,\max}\ \cdots\ f_{m,\max}]^T$ denotes the metrics' specifications and are given in advance. This assumption is quite reasonable since modeling PF at large metrics is useless because as assumed before a good design should have small metrics' values. As illustrated in Fig. 1, PF is part of the boundary of the feasible metric space $\Theta_m$. For those points located on PF (e.g., the red dots in Fig. 1), we cannot improve any one of the metrics without damaging the others. Note that in this paper, we only consider *non-degenerate* [20] and *convex* [12] PF, i.e., PF is a hypersurface with dimension $\Re^{m-1}$ and its convex hull is contained in $\Theta_m$, as it is always the case in real applications [4], [11]-[12].

Our purpose is to design an efficient PF modeling algorithm that can: (1) provide an accurate and generative model for PF, i.e., we can describe the PF with many generated points via the algorithm, and (2) fast examine whether a new provided metric vector $\mathbf{f}_{new} = \mathbf{f}(\mathbf{x}_{new})$ locates on PF or not. To keep the notation uncluttered, when there is no confusion we omit the dependence of $\mathbf{f}(\mathbf{x})$ on $\mathbf{x}$ from now on. Formally speaking, these two aforementioned requirements are equivalent to find a function mapping $h(\mathbf{f})$: $\Re^m \to \Re^1$, satisfying: (1) PF is given by the roots of $h(\bullet)$, i.e., $h(\mathbf{f}) = 0$, and all the roots are easily to generate, and (2) for a specific provided $\mathbf{f}_{new}$, we can easily evaluate $h(\mathbf{f}_{new})$.

### B. Relations to Existing Problems

It seems that the formulated problem can be solved by combining the multi-objective optimization method with state-of-the-art Data Mining method, regression, hypersurface fitting or classification. However, it is not the case, and in this sub-section we will discuss the relations of the formulated problems with existing Data Mining problems.

**Regression/Hypersurface fitting**. First, we adopt some multi-objective optimization method to obtain a subset of PF. Next, observing that PF is a hypersurface in $\Re^m$, we can use the obtained subset to train a regression model to fit this hypersurface.

For instance, we can define a non-linear regression model $h(\mathbf{f}) = \mathbf{w}^T \phi(\mathbf{f})$, where $\mathbf{w}$ is coefficient and $\phi(\bullet)$ is a selected basis function. Next, we use the obtained subset via multi-objective optimization method to fit the coefficients $\mathbf{w}$. However, this strategy cannot generate the whole PF since the roots of the non-linear function $h(\bullet)$ is non-trivial to obtain. Alternatively, if a simple linear function is chosen (e.g., $h(\mathbf{f}) = \mathbf{w}^T \mathbf{f}$), it will be easy to calculate the roots, but the generated PF will be far from accurate.

**Classification**. Similar to the previous case, after obtaining the subset of PF by solving multi-objective optimization, we can regard it as training set to learn a discriminative or generative model. To be more specific, we aim to model the joint distribution $pdf(A, \mathbf{f})$, (i.e., generative model) or the conditional distribution $pdf(A|\mathbf{f})$, (i.e., discriminative model) with the obtained training set, where A represents the event that $\mathbf{f}$ locates on PF. However, in this way, it is still difficult, if not impossible, to generate the PF. For instance, in this case, we regard the $\mathbf{f}$ satisfying $pdf(A|\mathbf{f}) \geq 0.5$ as a PF point. This will require us to traverse through the domain of $\mathbf{f}$ and search for $\mathbf{f}$ satisfying the requirement, and thus it will be time consuming.

Take generative adversarial network [21] (GAN) as an example. GAN is one of the most widely used generative model originally proposed for unsupervised learning, and it mainly consists of two parts: generator and discriminator. In our case, generator takes a random Gaussian noise as input and outputs a fake PF point, and the discriminator aims to distinguish the fake generated PF points from the true ones provided in the training set. Therefore, generator can be used to generate PF, and discriminator can be used to examine whether a new provided point locates on the true PF. Since PF is a small hypersurface in $\Re^m$ dimension, we may need to repeatedly iterate through generator and discriminator in order to generate the whole PF. For instance, among $10^3$ fake PF points generated by the generator, only 10 points may be judged on the true PF by the discriminator. Therefore, in order to describe the PF with $10^3$ points, we may need to iterate $10^5$ times. Although generating PF now seems feasible, it will be time unaffordable.

In summary, to the best of our knowledge, all state-of-the-art Data Mining methods, directly combined with multi-objective optimization, are not suitable in our problem. Moreover, in regression or classification, the training set has been assumed not computational expensive to obtain, and the main cost lies in model fitting. However, in our case, the training set is made up of PF points and obtaining those PF points via multi-objective optimization is already time consuming. Therefore, a wise algorithm should combine the step of obtaining PF points and the step of learning function $h(\bullet)$, so that it can decide how to query PF points in order to learn $h(\bullet)$ with as few samples as possible. Motivated by these observations, we develop a novel projection based active Gaussian process regression (P-aGPR) method for efficient PF modeling in the next Section.

## III. PROPOSED APPROACH

### A. Projection based Gaussian Process Regression

To start with, we consider the relation between the first $i$ metrics' PF and the projection of $m$ metrics' PF onto $\Re^i$, where $i = 1, 2, \cdots, m$. We use $\Gamma_i \subseteq \Re^i$ and $\Theta_i \subseteq \Re^i$ to denote PF and feasible metric space respectively when considering only the first $i$ metrics $\mathbf{f}_{1:i}(\mathbf{x}) = [f_1(\mathbf{x}) \ f_2(\mathbf{x}) \cdots f_i(\mathbf{x})]^T$. Moreover, we also define a projection function $\mathbf{g}_i(\mathbf{f}_{1:j})$, where $j = i, i+1, \cdots, m$, mapping the metric vector $\mathbf{f}_{1:j} = [f_1 \ f_2 \cdots f_i \cdots f_j]^T$ from $\Re^j$ to $\Re^i$ by maintaining the first $i$ entries in $\mathbf{f}_{1:j}$, i.e.,

$$\mathbf{g}_i(\mathbf{f}_{1:j}) = [f_1 \ f_2 \ \cdots \ f_i]^T, \quad (4)$$

and when $\mathbf{g}_i(\bullet)$ is performed on a set A made up of metrics vectors, it yields a set made up of all the metric vectors' projections, i.e.,

$$\mathbf{g}_i(A) = \{\mathbf{g}_i(\mathbf{f}_{1:j}) | \forall \mathbf{f}_{1:j} \in A\}. \quad (5)$$

Furthermore, we define a unary set operator $\Psi(A): A \rightarrow S$. Namely, when $\Psi(\bullet)$ is performed on a set A made up of metric vectors $\mathbf{f}_{1:j}$, it will form a subset $S \subseteq A$. For those metric vectors in S, they are non-dominated over each other, and conversely for some metric vector $\mathbf{f}_{1:j}$ in A but not in S (i.e., $\mathbf{f}_{1:j} \in A - S$), there must exist a metric vector $\mathbf{f}_{1:j}' \in S$ dominating $\mathbf{f}_{1:j}$. Intuitively, $\Psi(A)$ can be viewed as "extracting the PF of a given set A". For instance, based on definitions of $\Gamma_i$ and $\Theta_i$, we can write down:

$$\Psi(\Theta_i) = \Gamma_i, i = 1, 2, \cdots, m. \quad (6)$$

After introducing these notations, we observe the following four important properties: (i) $\mathbf{g}_i(\Theta_m) = \Theta_i$. The projection of feasible metrics space $\mathbf{g}_i(\Theta_m)$ is identical to the feasible metrics space $\Theta_i$. (ii) $\Gamma_i = \Psi(\Theta_i) \subseteq \Theta_i$, which is intuitive and shown in Eq (6). (iii) $\Gamma_i = \Psi(\mathbf{g}_i(\Gamma_m))$. Performing $\Psi(\bullet)$ on the projection $\mathbf{g}_i(\Gamma_m)$ is identical to $\Gamma_i$. (iv) If $\mathbf{f}_{1:i} = \mathbf{g}_i(\mathbf{f}_{1:m}) \notin \mathbf{g}_i(\Gamma_m)$, we must have $\mathbf{f}_{1:i+1} = \mathbf{g}_{i+1}(\mathbf{f}_{1:m}) \notin \mathbf{g}_{i+1}(\Gamma_m)$. Except property (iii), the other three properties are quite obvious, and we have provided a proof for (iii) in Appendix.

The aforementioned four observations motivate our proposed P-aGPR method. Namely, we can iterate from $i = 1$ to $i = m$, and examine whether $\mathbf{f}_{1:i}$ is contained in $\mathbf{g}_i(\Gamma_m)$. If not, we can immediately stop and assert that $\mathbf{f}_{1:m}$ doesn't locate on PF $\Gamma_m$, as shown in Fig. 2(a). As illustrated later, besides examining whether a point locates on PF, this algorithm flow can equally be applied to generatively model PF. Now we need to find mathematical expression for each condition in the square bracket of Fig. 2(a).

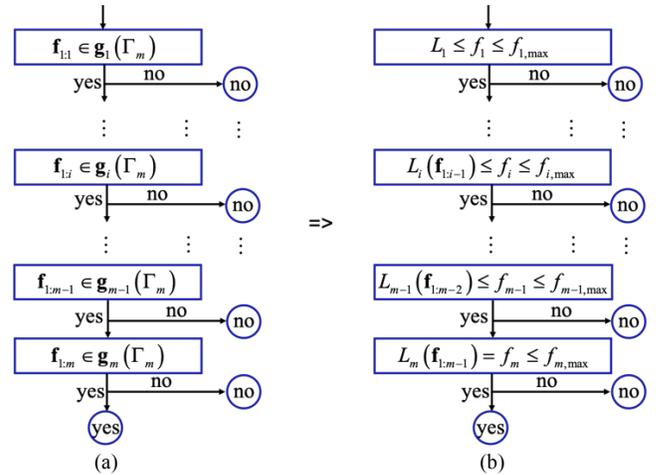

Fig. 2. (a) Illustration of algorithm key idea. We examine the condition in each square bracket from top to bottom. The "no" circle represents that $\mathbf{f}_{1:m}$ doesn't locate on Pareto Front $\Gamma_m$, and on the other hand "yes" circle represents that $\mathbf{f}_{1:m}$ locates on Pareto Front $\Gamma_m$. (b) algorithm flow expressed in several inequalities and one equality.

Let's use a simple 3-dimensional example (as shown in Fig. 3) to illustrate the key idea. Fig. 3(a) shows a PF in 3-dimensional space (i.e., $m = 3$), and $\Gamma_3$ is the shaded quarter sphere. Suppose we want to model $\Gamma_3$ in the region satisfying $\mathbf{f}_{1:3} \leq \mathbf{f}_{max} = [f_{1,max} \ f_{2,max} \ f_{3,max}]^T = [1 \ 1 \ 1]^T$. As shown in Fig. 3(a), $\Gamma_3$ seems well expressed by a non-linear function $f_3 = L_3(f_1, f_2)$. However, as shown in Fig. 3(b), modeling $\Gamma_3$ merely with $f_3 = L_3(f_1, f_2)$ is not enough, since $[f_1 \ f_2]^T$ can only locate in the shaded quarter circle and this constraint is not captured by $L_3(f_1, f_2)$. Therefore, in this case, to accurately model $\Gamma_3$, we need three functions $\{L_1, L_2, L_3\}$ in total:

$$\begin{aligned} L_1 &\leq f_1 \leq f_{1,\max} \\ L_2(f_1) &\leq f_2 \leq f_{2,\max} \\ L_3(f_1, f_2) &= f_3 \leq f_{3,\max} \end{aligned} \quad (7)$$

Note that $L_1$ is a constant function, and $\{L_1, L_2, L_3\}$ represents the constraints for $\{f_1, f_2, f_3\}$ respectively. Compared Eq (7) to Fig. 2(a), we extend to a general $m$-dimensional case. Namely, in Fig. 2(a), we can equivalently write the condition in the $i$-th square bracket as:

$$\begin{aligned} L_1 &\leq f_1 \leq f_{1,\max}, i = 1 \\ L_i(\mathbf{f}_{1:i-1}) &\leq f_i \leq f_{i,\max}, i = 2, 3, \cdots, m-1. \\ L_m(\mathbf{f}_{1:m-1}) &= f_m \leq f_{m,\max}, i = m \end{aligned} \quad (8)$$

In other words, if we can fit $\{L_i; i = 1, 2, \cdots, m\}$ provided some training samples, we can easily examine whether a new provided metric vector $\mathbf{f}_{1:m}$ locates on $\Gamma_m$ or not as shown in Fig. 2(b). Moreover, this algorithm flow can also be used to model $\Gamma_m$ generatively. To be more specific, we first randomly generate $f_1$ in interval $[L_1, f_{1,\max}]$, and next we randomly generate $f_2$ in interval $[L_2(f_1), f_{2,\max}]$ using the generated $f_1$. This procedure is repeated until all entries of $\mathbf{f}_{1:m}$ have been generated. Moreover, unlike GAN discussed in Section II.B, we can guarantee that each generated point is a PF point.

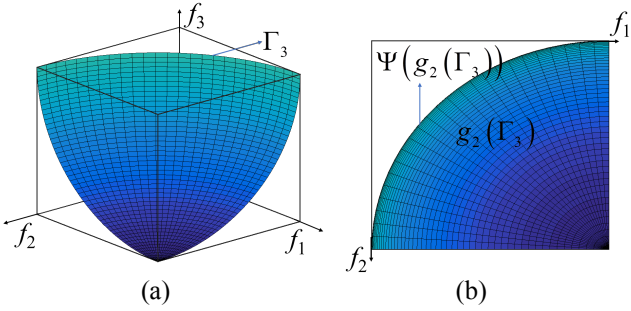

Fig. 3. (a) a 3-dimensional PF example. The Pareto Front $\Gamma_3$ is the shaded quarter sphere, which is the surface of a unit ball centering at $[1\ 1\ 1]^T$ with radius equal to 1. $\Theta_3$ is the space made up of extending $\Gamma_3$ to the up-right position. (b) Only the first two metrics $f_1$ and $f_2$ are considered. $\Psi(\mathbf{g}_2(\Gamma_3))$ is an arc, and $\Theta_2$ is the space made up of extending $\Psi(\mathbf{g}_2(\Gamma_3))$ to the left-down position. The shaded quarter circle is $\mathbf{g}_2(\Gamma_3)$. Here in this case, $\mathbf{g}_2(\Gamma_3) \subseteq \Theta_2$.

The remaining problem is how to obtain the training sets $\{D_i; i = 1, 2, \cdots, m\}$, and next how to use them to fit $\{L_i; i = 1, 2, \cdots, m\}$. In this paper, we propose to adopt a modified NBI method [11]-[12] to obtain the training sets and Gaussian process regression (GPR) models to learn $\{L_i; i = 1, 2, \cdots, m\}$. Recall that in Fig. 3(b), $L_2(f_1)$ actually represents the arc $\Psi(\mathbf{g}_2(\Gamma_3))$, which is identical to $\Gamma_2$ as stated in property (iii). Namely, instead of finding points in $\Psi(\mathbf{g}_2(\Gamma_3))$, we can directly find points in $\Gamma_2$ to construct a training set $D_2$. This will reduce the computational cost since $\Psi(\mathbf{g}_2(\Gamma_3))$ requires calculating $\Gamma_3$, performing $\mathbf{g}_2(\bullet)$ and $\Psi(\bullet)$ in sequence, while on the other hand directly dealing with $\Gamma_2$ is much easier. Extending to other dimensions, if we can find a data set $D_k$ made up of metric vectors $\mathbf{f}_{1:k}$ contained in $\Gamma_k$, we can use $D_k$ to train a GPR model to learn $L_k(\bullet)$, where $k = 2, 3, \cdots, m$. Note that the constant function $L_1$ is simply given by the minimum value that $f_1$ can achieve, so that there is no need to obtain $D_1$ and fit a GPR model for $L_1$. Now the remaining problem is how to obtain the training sets $\{D_k; k = 2, 3, \cdots, m\}$.

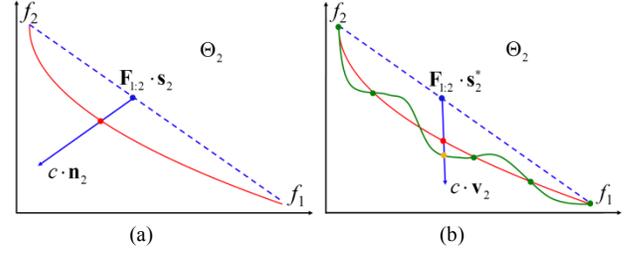

Fig. 4. (a) Illustration of NBI method in 2-dimensional space. The red curve represents $\Gamma_2$ and the blue dashed line represents the convex hull of the columns of $\mathbf{F}_{1:2}$. The red dot deontes the found PF point. (b) Illustration of querying a new data point vertically in active learning framework. The green dots denote the current $D_2$, and the green curve represents the learned $\mu_2(\mathbf{f}_{1:1})$ by $GPR_2$. The yellow, blue and red dot represents $\mathbf{f}_{1:2}^{(\text{around})}$, $\mathbf{F}_{1:2} \cdot \mathbf{s}_2^*$, and the next found PF point $\mathbf{f}_{1:2}(\mathbf{x}^{\text{opt}})$ respectively.

In this paper, we adopt a modified NBI method [11]-[12] to obtain $\{D_k; k = 2, 3, \cdots, m\}$. NBI is originally proposed in [12] to solve multi-objective optimization problem with convex PF. We start from solving $m$ single-objective optimization problems separately:

$$\mathbf{x}^{i,\text{opt}} = \arg\min_{\mathbf{x} \in \Omega} f_i(\mathbf{x}), i = 1, 2, \cdots, m, \quad (9)$$

and $f_{i,\min} = f_i(\mathbf{x}^{i,\text{opt}})$ represents the minimum value that the $i$-th metric can achieve. Note that we can directly set $L_1 = f_{1,\min}$. Furthermore, we collectively denote them as a vector $\mathbf{f}_{\min} = [f_{1,\min}\ f_{2,\min}\ \cdots\ f_{m,\min}]^T$. Without loss of generality, we assume that all entries of an arbitrary metric vectors $\mathbf{f}_{1:m}$ in $\Theta_m$ are no less than 0 (i.e., $\mathbf{f}_{1:m} \geq \mathbf{0}$). If some metric vector $\mathbf{f}_{1:m}$ violates this constraint, we can re-define the metric vector as $\mathbf{f}_{1:m} - \mathbf{f}_{\min}$. Next, we can initialize a matrix $\mathbf{F}$ as follows:

$$\mathbf{F} = \begin{bmatrix} f_{1,\min} & f_{1,\max} & \cdots & f_{1,\max} \\ f_{2,\max} & f_{2,\min} & \cdots & f_{2,\max} \\ \vdots & \vdots & \vdots & \vdots \\ f_{m,\max} & f_{m,\max} & \cdots & f_{m,\min} \end{bmatrix}, \quad (10)$$

and we can solve the following scalarized optimization problem to yield one sample in $D_k$:

$$\begin{aligned} \max_{c \geq 0, \mathbf{x} \in \Omega} \quad & c \\ s.t. \quad & \mathbf{F}_{1:k} \cdot \mathbf{s}_k + c \cdot \mathbf{n}_k = \mathbf{f}_{1:k}(\mathbf{x}), \\ & \mathbf{s}_k \geq \mathbf{0}, \mathbf{s}_k^T \cdot \mathbf{e}_k = 1 \end{aligned} \quad (11)$$

where $\mathbf{e}_k \in \mathfrak{R}^k$ is a vector with all entries equal to 1, $\mathbf{F}_{1:k} \in \mathfrak{R}^{k \times k}$ is the sub-matrix made up of the first $k$ rows and $k$ columns of $\mathbf{F}$. To be more specific, $\mathbf{s}_k \in \mathfrak{R}^k$ is a weight vector with all entries no less than 0 and the summation of its entries equals 1, and $\mathbf{n}_k \in \mathfrak{R}^k$ denotes a pre-defined search direction, e.g., $\mathbf{n} = -\mathbf{F}_{1:k} \cdot \mathbf{e}_k$ [11], etc. Namely, we start from a point $\mathbf{F}_{1:k} \cdot \mathbf{s}_k$ and extend along the direction defined by $\mathbf{n}_k$. We aim to find the maximum length, represented by $c \cdot \mathbf{n}_k$, enforcing the end point is contained in the feasible metric space $\Theta_k$. The intrinsic idea, illustrated in Fig. 4(a), is that the convex PF locates "beneath" the convex hull of the columns of $\mathbf{F}_{1:k}$ [11].

Solving Eq (11) will yield an optimal $(c^{\text{opt}}, \mathbf{x}^{\text{opt}})$ pair, and $\mathbf{f}_{1:k}(\mathbf{x}^{\text{opt}})$ can be added to data set $D_k$ in the form of $\{\mathbf{f}_{1:k-1}(\mathbf{x}^{\text{opt}}),$

$t_k = f_k(\mathbf{x}^{\text{opt}})\}$. By repeatedly altering $\mathbf{s}_k$ and next solving Eq (11), we can obtain the required data set $D_k = \{(\mathbf{f}_{1:k-1}^{(n)}, t_k^{(n)}); n = 1, 2, \cdots, N\}$, where $N$ is the number of samples in $D_k$. Note that we assume that the number of samples in every training set $\{D_k; k = 2, 3, \cdots, m\}$ equals $N$.

With the training sets $\{D_k; k = 2, 3, \cdots, m\}$, we build $(m-1)$ GPR models for $\{L_k; k = 2, 3, \cdots, m\}$, denoted as $\{\text{GPR}_k; k = 2, 3, \cdots, m\}$. $\text{GPR}_k$ can be fully characterized by a mean function $\rho_k(\mathbf{f}_{1:k-1})$ and a covariance function $\kappa_k(\mathbf{f}_{1:k-1}^{(p)}, \mathbf{f}_{1:k-1}^{(q)})$ [14]. In this paper, we choose a constant zero mean function and the squared exponential covariance function:

$$\rho_k(\mathbf{f}_{1:k-1}) = 0$$
$$\kappa_k\left(\mathbf{f}_{1:k-1}^{(p)}, \mathbf{f}_{1:k-1}^{(q)}\right) = \theta_{k,1} \cdot \exp\left[-\frac{\theta_{k,2}}{2}\left\|\mathbf{f}_{1:k-1}^{(p)} - \mathbf{f}_{1:k-1}^{(q)}\right\|^2\right], \quad (12)$$

where $\{\theta_{k,1}, \theta_{k,2}\}$ are hyper-parameters for $\text{GPR}_k$. Namely, trained with the data set $D_k = \{(\mathbf{f}_{1:k-1}^{(n)}, t_k^{(n)}); n = 1, 2, \cdots, N\}$, $\text{GPR}_k$ can give a probabilistic prediction for $L_k(\mathbf{f}_{1:k-1})$, i.e.,

$$L_k(\mathbf{f}_{1:k-1}) | D_k \sim N\left(\mu_k(\mathbf{f}_{1:k-1}), \sigma_k^2(\mathbf{f}_{1:k-1})\right), \quad (13)$$

where the mean $\mu_k(\mathbf{f}_{1:k-1})$ and variance $\sigma_k^2(\mathbf{f}_{1:k-1})$ both depend on $\mathbf{f}_{1:k-1}$, and $\mu_k(\mathbf{f}_{1:k-1})$ is given by [14]:

$$\mu_k(\mathbf{f}_{1:k-1}) = \boldsymbol{\kappa}_k(\mathbf{f}_{1:k-1}, D_k)^T \cdot \mathbf{K}_k^{-1} \cdot \mathbf{t}_k, \quad (14)$$

where $\mathbf{t}_k \in \Re^N$ is a column vector defined as $\mathbf{t}_k = [t_k^{(1)} \, t_k^{(2)} \cdots t_k^{(N)}]^T$, and $\boldsymbol{\kappa}_k(\mathbf{f}_{1:k-1}, D_k) = [\kappa_k(\mathbf{f}_{1:k-1}, \mathbf{f}_{1:k-1}^{(1)}) \, \kappa_k(\mathbf{f}_{1:k-1}, \mathbf{f}_{1:k-1}^{(2)}) \cdots \kappa_k(\mathbf{f}_{1:k-1}, \mathbf{f}_{1:k-1}^{(N)})]^T \in \Re^N$ represents the correlation between $\mathbf{f}_{1:k-1}$ with the data set $D_k$. $\mathbf{K}_k$ is a $\Re^{N \times N}$ matrix whose $(i, j)$-th entry is given by $\kappa_k(\mathbf{f}_{1:k-1}^{(i)}, \mathbf{f}_{1:k-1}^{(j)})$. The variance $\sigma_k^2(\mathbf{f}_{1:k-1})$ in Eq (13) is given by [14]:

$$\sigma_k^2(\mathbf{f}_{1:k-1}) = \kappa_k(\mathbf{f}_{1:k-1}, \mathbf{f}_{1:k-1}) - \boldsymbol{\kappa}_k(\mathbf{f}_{1:k-1}, D_k)^T \cdot \mathbf{K}_k^{-1} \cdot \boldsymbol{\kappa}_k(\mathbf{f}_{1:k-1}, D_k). \quad (15)$$

Algorithm 1 summarizes the main steps of initialization of the proposed projection based GPR models. After obtaining $L_1$ and learning all $(m-1)$ GPR models for $\{L_k; k = 2, 3, \cdots, m\}$, we can follow the algorithm flow in Fig. 2(b) from top to bottom, in order to either model the PF generatively or check whether a new provided metric vector locates on PF.

**Algorithm 1: Initialization of Projection GPR**
1. Give metrics' specifications $\mathbf{f}_{\max} = [f_{1,\max} f_{2,\max} \cdots f_{m,\max}]^T$.
2. Solve $m$ single-objective optimizations as in Eq (9).
3. Set $L_1 = f_{1,\min}$ and initialize matrix $\mathbf{F}$ as in Eq (10).
4. Define all hyper-parameters $\{(\theta_{k,1}, \theta_{k,2}); k = 2, 3, \cdots, m\}$.
5. For $k = 2:m$
6.    Solve Eq (11) with $\mathbf{n}_k = -\mathbf{F}_{1:k} \cdot \mathbf{e}_k$ for $N$ times, and at each time randomly choose a weight vector $\mathbf{s}_k$.
7.    Obtain data set $D_k = \{(\mathbf{f}_{1:k-1}^{(n)}, t_k^{(n)}); n = 1, 2, \cdots, N\}$.
8.    Fit $\text{GPR}_k$ for $L_k$ according to Eq (13)-(15).
9. EndFor

*B. Active Learning*

Suppose now $N_{\max}$ samples are allowed for training each $L_k$. The most trivial implementation is that we can simply set $N = N_{\max}$ in Algorithm 1. However, this is not efficient since the uncertainty information of GPR models is completely ignored. To be more specific, GPR model can give the uncertainty about the prediction at a new point according to Eq (15). Utilizing this uncertainty information, we may decide where to query the samples, so that we can train the GPR models with as few samples as possible.

This observation motivates us to adopt the *active learning* [15]-[19] framework to fully exploit the uncertainty information. Active learning is a special framework allowing the learning algorithm to query the data from the knowledge that has already been obtained. It is especially suitable when obtaining a training sample is computational expensive as in our case. First, we adopt Algorithm 1 to initialize those $(m-1)$ GPR models by setting $N = N_0$ ($N_0 < N_{\max}$). Next, we iterate through each GPR model and for $\text{GPR}_k$ (where $k = 2, 3, \cdots, m$), we can find the position with the maximum variance, i.e., we solve the following problem:

$$\max_{\mathbf{f}_{1:k-1}} \sigma_k^2(\mathbf{f}_{1:k-1})$$
$$s.t. \quad L_1 \leq f_1 \leq f_{1,\max} \quad , \quad (16)$$
$$\mu_j(\mathbf{f}_{1:j-1}) \leq f_j \leq f_{j,\max}, j = 2, 3, \cdots, k-1$$

where the inequality constraints make sure that the optimal solution of Eq (16), denoted as $\mathbf{f}_{1:k-1}^{(\text{query})}$, is contained in $\mathbf{g}_{k-1}(\Gamma_m)$. Note that when the initial number of training samples $N_0$ is small, the GPR models are not accurate enough, and thus the inequality constraints may not suffice to guarantee $\mathbf{f}_{1:k-1}^{(\text{query})} \in \mathbf{g}_{k-1}(\Gamma_m)$. This will be discussed later in Section IV. Next exploiting $\text{GPR}_k$, we can obtain the predictive value of $L_k(\mathbf{f}_{1:k-1}^{(\text{query})})$, i.e., $\mu_k(\mathbf{f}_{1:k-1}^{(\text{query})})$. In summary, we know that $\text{GPR}_k$ has maximum variance at $\mathbf{f}_{1:k-1}^{(\text{query})}$, and the current prediction at $\mathbf{f}_{1:k-1}^{(\text{query})}$ equals $\mu_k(\mathbf{f}_{1:k-1}^{(\text{query})})$.

Now we adopt NBI method to find the true PF point in $\Gamma_k$ with the first $(k-1)$ entries equal $\mathbf{f}_{1:k-1}^{(\text{query})}$, and this PF point will be added to $D_k$ to retrain $\text{GPR}_k$. In other words, the next found PF point represents where $\text{GPR}_k$ is most uncertain about. According to Eq (11), when search direction $\mathbf{n}_k$ is chosen, one PF point solution corresponds to a specific weight vector $\mathbf{s}_k$. Therefore, if we want to find a point with first $(k-1)$ entries set to $\mathbf{f}_{1:k-1}^{(\text{query})}$, we first need to set $\mathbf{n}_k$ as a vertical vector $\mathbf{v}_k = [0 \, 0 \cdots 0 \, -1]^T \in \Re^k$, and next determine the weight vector $\mathbf{s}_k$ by solving:

$$\mathbf{F}_{1:k} \cdot \mathbf{s}_k + c \cdot \mathbf{v}_k = \mathbf{f}_{1:k}^{(\text{around})}, \quad (17)$$
$$s.t. \quad \mathbf{s}_k \geq \mathbf{0}, \mathbf{s}_k^T \cdot \mathbf{e}_k = 1$$

where $\mathbf{f}_{1:k}^{(\text{around})} = [\mathbf{f}_{1:k-1}^{(\text{query})}; \mu_k(\mathbf{f}_{1:k-1}^{(\text{query})})] \in \Re^k$. Eq (17) can be solved analytically if neglecting the constraint $\mathbf{s}_k \geq \mathbf{0}$:

$$c^* = \frac{\left(\mathbf{F}_{1:k}^{-1} \cdot \mathbf{f}_{1:k}^{(\text{around})}\right)^T \cdot \mathbf{e}_k - 1}{\left(\mathbf{F}_{1:k}^{-1} \cdot \mathbf{v}_k\right)^T \cdot \mathbf{e}_k}. \quad (18)$$
$$\mathbf{s}_k^* = \mathbf{F}_{1:k}^{-1} \cdot \left(\mathbf{f}_{1:k}^{(\text{around})} - c^* \cdot \mathbf{v}_k\right)$$

Note that if $\mathbf{f}_{1:k}^{(\text{around})}$ is close to a true PF point in $\Gamma_k$, the constraint $\mathbf{s}_k \geq \mathbf{0}$ will be satisfied. We will discuss in next Section if some entry of $\mathbf{s}_k^*$ is less than 0. Next, using the obtained weight vector $\mathbf{s}_k^*$, we solve the NBI problem:

$$\max_{c \geq 0, \mathbf{x} \in \Omega} c$$
$$s.t. \quad \mathbf{F}_{1:k} \cdot \mathbf{s}_k^* + c \cdot \mathbf{v}_k = \mathbf{f}_{1:k}(\mathbf{x}) \quad . \quad (19)$$

As mentioned before, solving Eq (19) will yield an optimal $(c^{\text{opt}}, \mathbf{x}^{\text{opt}})$ pair, and $\mathbf{f}_{1:k}(\mathbf{x}^{\text{opt}})$ can be used to update data set $D_k$, i.e., $D_k = D_k \cup \{\mathbf{f}_{1:k-1}(\mathbf{x}^{\text{opt}}), t_k = f_k(\mathbf{x}^{\text{opt}})\}$. Note that here ideally we have $\mathbf{f}_{1:k-1}(\mathbf{x}^{\text{opt}}) = \mathbf{f}_{1:k-1}^{(\text{query})}$ since we have chosen $\mathbf{n}_k = \mathbf{v}_k$ and fix weight vector $\mathbf{s}_k$ to $\mathbf{s}_k^*$ in advance. This modified NBI method with vertical search direction has been illustrated in Fig. 4(b).

After updating the training set $D_k$, we can retrain $GPR_k$ with it. Next, we still query a PF point with maximum variance based on the updated $GPR_k$. This procedure will be repeated until the size of $D_k$ reaches $N_{max}$. Then we move to the next data set $D_{k+1}$ and repeat the two steps, i.e., querying a PF point with maximum variance and updating the GPR model. We will iterate through $\{D_k; k = 2, 3, \cdots, m\}$ until the sizes of all data sets reach $N_{max}$.

## IV. IMPLEMENTATION DETAILS

The practicability of the NBI method relies on that the weight vector $\mathbf{s}_k$ should be no less than $\mathbf{0}$, and that the summation of its entries equals 1. However, in our current proposed method, the analytical solution Eq (18) only guarantees that the summation equals 1, and we are not sure whether $\mathbf{s}_k^* \geq \mathbf{0}$. In this Section we will discuss several important implementation details addressing this issue.

### A. Rectified NBI Method

There are two possible reasons which may lead to some entry of $\mathbf{s}_k^*$ less than 0. The first one is that vertical search direction $\mathbf{n}_k = \mathbf{v}_k$ may not be sufficed to cover all PF points. As illustrated in Fig. 5, when weight vector $\mathbf{s}_3$ satisfies $\mathbf{s}_3 \geq \mathbf{0}$ and $\mathbf{s}_3^T \cdot \mathbf{e}_3 = 1$, $\mathbf{F}_{1:3} \cdot \mathbf{s}_3$ must lie in the red triangle with vertex given by $\mathbf{o}_1$, $\mathbf{o}_2$ and $\mathbf{o}_3$. Therefore, vertical search in this case cannot find those PF points "inside" the brown plane, i.e., we cannot find PF points satisfying $f_1 + f_2 < 1$. To address this issue, notice that if we choose the search direction as $\mathbf{n} = -\mathbf{F}_{1:3} \cdot \mathbf{e}_3 = [-1\ -1\ -1]^T$, we can cover all the PF points. Therefore, we can first set $\mathbf{n}_k = \mathbf{v}_k$, and if some entry of the corresponding $\mathbf{s}_k^*$ solved in Eq (17)-(18) is less than 0, we reset $\mathbf{n}_k = -\mathbf{F}_{1:k} \cdot \mathbf{e}_k$ and adopt NBI method to find a PF point along this search direction again. To be more specific, when some entry of $\mathbf{s}_k^*$ solved in Eq (17)-(18) is less than 0, we revise Eq (17) as:

$$\mathbf{F}_{1:k} \cdot \mathbf{s}_k + c \cdot \left(-\mathbf{F}_{1:k} \cdot \mathbf{e}_k\right) = \mathbf{f}_{1:k}^{(around)} \quad (20)$$
$$s.t. \quad \mathbf{s}_k \geq \mathbf{0}, \mathbf{s}_k^T \cdot \mathbf{e}_k = 1$$

Similar to Eq (18), it can be solved analytically:

$$c^{*'} = \frac{1 - \left(\mathbf{F}_{1:k}^{-1} \cdot \mathbf{f}_{1:k}^{(around)}\right)^T \cdot \mathbf{e}_k}{k}, \quad (21)$$

$$\mathbf{s}_k^{*'} = \mathbf{F}_{1:k}^{-1} \cdot \mathbf{f}_{1:k}^{(around)} + c^{*'} \cdot \mathbf{e}_k$$

and next we adopt NBI method along $\mathbf{n}_k = -\mathbf{F}_{1:k} \cdot \mathbf{e}_k$, i.e.,

$$\max_{c \geq 0, \mathbf{x} \in \Omega} c \quad (22)$$
$$s.t. \quad \mathbf{F}_{1:k} \cdot \mathbf{s}_k^{*'} + c \cdot \left(-\mathbf{F}_{1:k} \cdot \mathbf{e}_k\right) = \mathbf{f}_{1:k}(\mathbf{x})$$

One important thing worthy mentioned is that first setting $\mathbf{n}_k = \mathbf{v}_k$ is necessary, since this will make sure that the first $(k-1)$ entries of the next found PF points equal $\mathbf{f}_{1:k-1}^{(query)}$, and thus the next found PF point will correspond to where $GPR_k$ is most uncertain about. Only when some entry of the $\mathbf{s}_k^*$ solved in Eq (17)-(18) is less than 0, we will reset $\mathbf{n}_k = -\mathbf{F}_{1:k} \cdot \mathbf{e}_k$ and adopt Eq (20)-(22).

The second reason leading some entry of $\mathbf{s}_k^*$ less than 0 is that the GPR models are not accurate, which cannot be remedied by resetting $\mathbf{n}_k = -\mathbf{F}_{1:k} \cdot \mathbf{e}_k$, i.e., some entry of $\mathbf{s}_k^{*'}$ solved in Eq (21) is still negative. To be more specific, when GPR models are not accurate enough, the optimal solution $\mathbf{f}_{1:k-1}^{(query)}$ obtained by Eq (16) may not be contained in $\mathbf{g}_{k-1}(\Gamma_m)$. Moreover, if the GPR models are not accurate, the current prediction of $L_k(\mathbf{f}_{1:k-1}^{(query)})$, i.e., $\mu_k(\mathbf{f}_{1:k-1}^{(query)})$, will also be far from its true value, and thus $\mathbf{f}_{1:k}^{(around)} = [\mathbf{f}_{1:k-1}^{(query)}; \mu_k(\mathbf{f}_{1:k-1}^{(query)})]$ will also be far from $\Gamma_k$. In this case, no matter setting $\mathbf{n}_k = \mathbf{v}_k$ or $\mathbf{n}_k = -\mathbf{F}_{1:k} \cdot \mathbf{e}_k$, the corresponding $\mathbf{s}_k^*$ or $\mathbf{s}_k^{*'}$ solved by Eq (17)-(18) or Eq (20)-(21) may both have entry less than 0 due to the error of $\mathbf{f}_{1:k}^{(around)}$. To address this issue, we adopt a simple heuristic modification to $\mathbf{s}_k^{*'}$. Namely, when $\mathbf{s}_k^{*'}$ still has entry less than 0, we reset all these negative entries to 0 and renormalize $\mathbf{s}_k^{*'}$ to enforce the summation of its entries equal to 1. Then we solve Eq (22) to obtain a PF point with this renormalized $\mathbf{s}_k^{*'}$.

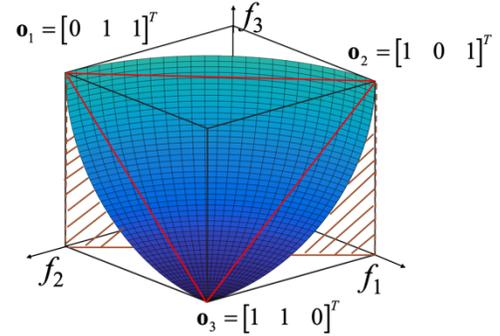

Fig. 5. Illustration of the fact that vertical search may not cover all PF points. The Pareto Front $\Gamma_3$ is the blue shaded quarter sphere as in Fig. 3(a). Here $\mathbf{o}_1$, $\mathbf{o}_2$ and $\mathbf{o}_3$ are the three columns in matrix $\mathbf{F}_{1:3}$.

### B. Summary

Algorithm 2 summarizes the proposed projection based active GPR (P-aGPR) method for PF modeling. From Step 1 to Step 3, we first adopt Algorithm 1 to initialize all GPR models with $N_0$ samples allowed for each. Next, we iterate through $\{GPR_k; k = 2, 3, \cdots, m\}$, and for $GPR_k$ we query a new PF point with the maximum variance in Step 5. Then we adopt the rectified NBI method from Step 6 to Step 11, and update the data set $D_k$ correspondingly. Finally, we retrain $GPR_k$ with the updated $D_k$. We repeat querying a new PF point and retraining $GPR_k$ until the size of $D_k$ reaches $N_{max}$.

**Algorithm 2: Projection based Active GPR (P-aGPR)**
1. Give maximum allowed samples $N_{max}$ for training each $\{L_k; k = 2, 3, \cdots, m\}$.
2. Choose the number of initial samples $N_0 < N_{max}$.
3. Adopt Algorithm 1 with $N = N_0$.
4. For $k = 2:m$
5.    Solve Eq (16) to obtain maximum variance position $\mathbf{f}_{1:k-1}^{(query)}$.
6.    Solve Eq (17)-(18) yielding $\mathbf{s}_k^*$.
7.    If $\mathbf{s}_k^* \geq \mathbf{0}$, solve (19) and update $D_k$. Then go to step 12.
8.    Solve Eq (20)-(21) yielding $\mathbf{s}_k^{*'}$.
9.    If $\mathbf{s}_k^{*'} \geq \mathbf{0}$, solve (22) and update $D_k$. Then go to step 12.
10.   Reset all negative entries in $\mathbf{s}_k^{*'}$ to 0, and renormalize $\mathbf{s}_k^{*'}$ to enforce the summation of its entries equal to 1.
11.   Solve Eq (22) with the renormalized $\mathbf{s}_k^{*'}$ and update $D_k$.
12.   Retrain $GPR_k$ with $D_k$.

13. If the size of $D_k$ is still less than $N_{max}$, go back to step 5.
14. EndFor

## V. NUMERICAL EXAMPLES

In this Section, four testbench functions have been used to test the efficacy of the proposed P-aGPR method. For comparison purpose, two *passive* modeling methods have been implemented. All numerical experiments are performed on a computer cluster composed of 500 workstations and each workstation is equipped with a 2.67GHz CPU and 4GB memory.

### A. Experiment Setup

To the best of our knowledge, directly combining state-of-the-art Data Mining methods with multi-objective optimization cannot address our issue as discussed in Section II.B. Therefore, for comparison purpose, we implement three methods: (1) the proposed P-aGPR method, (2) projection based passive GPR (P-pGPR) method, in which we directly train all GPR models by adopting Algorithm 1 with $N = N_{max}$, (3) projection based passive polynomial regression (P-pPR) method, in which we use a polynomial to learn $L_k$ instead of a GPR model, i.e., $L_k(\mathbf{f}_{1:k-1}) = \mathbf{w}^T \boldsymbol{\phi}(\mathbf{f}_{1:k-1}) + \varepsilon$. Here $\mathbf{w}$ is the coefficient to learn, $\varepsilon$ is a Gaussian random noise, and $\boldsymbol{\phi}(\mathbf{f}_{1:k-1})$ is a polynomial basis function with all terms up to degree 2, e.g., $\boldsymbol{\phi}(\mathbf{f}_{1:2}) = 1 + f_1 + f_2 + f_1 \times f_2 + f_1^2 + f_2^2$. The proposed P-aGPR method is an active learning method while the other two can be regarded as passive learning methods (i.e., directly trained with randomly given samples) with different regression models.

TABLE I. TESTBENCH FUNCTIONS SUMMARY

| | Design variable dim $d$ | Metric dim $m$ | Metric spec $\mathbf{f}_{max}$ | Pareto Front (PF) |
|---|---|---|---|---|
| ZDT1[22] | 6 | 2 | $[1\ 1]^T$ | $f_1^{0.5} + f_2 = 1$ |
| SCH[23] | 1 | 2 | $[4\ 4]^T$ | $(f_1^{0.5} - 2)^2 = f_2$ |
| SPH[24] | 3 | 3 | $[0\ 0\ 0]^T$ | $f_1^2 + f_2^2 + f_3^2 = 1$ |
| MAF3[25] | 4 | 3 | $[0.25\ 0.25\ 1]^T$ | $f_1^{0.5} + f_2^{0.5} + f_3 = 1$ |

a. SPH function has been modified

Table I summarizes four testbench functions used in this paper. The first two ZDT1 and SCH are 2-dimensional testbenches (i.e., $m = 2$) while the last two SPH and MAF3 are 3-dimensional testbenches (i.e., $m = 3$). Note that SPH [24] has been modified in this paper, the original PF of the SPH function is given by $f_1^2 + f_2^2 + f_3^2 = 1$, $0 \le f_1 \le 1$, $0 \le f_2 \le 1$, $0 \le f_1 \le 1$. In this paper, we set all the metrics' values to their negatives, and thus the PF is given by $f_1^2 + f_2^2 + f_3^2 = 1$, $-1 \le f_1 \le 0$, $-1 \le f_2 \le 0$, $-1 \le f_1 \le 0$. To characterize the efficacy of the proposed P-aGPR method, we need to answer two main questions: (1) how well is the learned PF, and (2) how fast can we learn the PF?

To answer the second question, we can measure the CPU running time of different PF modeling methods. To answer the first question, we generate $N_{PF}$ points with the PF modeling method (i.e., P-aGPR, P-pGPR or P-pPR), where $N_{PF} = 1000$ when testbench is ZDT1 or SCH and $N_{PF} = 8000$ when testbench is SPH or MAF3. Then for each generated point, we calculate its minimum Euclidean distance to the data points on the true PF. To be more specific, we define an absolute distance error *Err*:

$$Err = \frac{1}{N_{PF}} \sum_{n=1}^{N_{PF}} dis\left(\mathbf{f}_{1:m}^{(n),\text{gen}}, \Gamma_m\right), \quad (23)$$

where $\mathbf{f}_{1:m}^{(n),\text{gen}}$ is the $n$-th generated PF point, and we define:

$$dis\left(\mathbf{f}_{1:m}^{(n),\text{gen}}, \Gamma_m\right) = \min_{\mathbf{f}_{1:m} \in \Gamma_m} \left\|\mathbf{f}_{1:m}^{(n),\text{gen}} - \mathbf{f}_{1:m}\right\|. \quad (24)$$

Namely, *Err* can be used to measure the accuracy of the generated PF with different methods.

For fairness, the hyper-parameters for GPRs in p-aGPR and P-pGPR method are identical. Moreover, finding a PF point requires to solve the NBI sub-problem as in Eq (11), (19) and (22). This non-linear optimization problem is addressed by an optimization solver and its parameters, e.g., convergence tolerance, step length, etc., are set identical for all three PF modeling methods.

### B. Experiment Results and Analyses

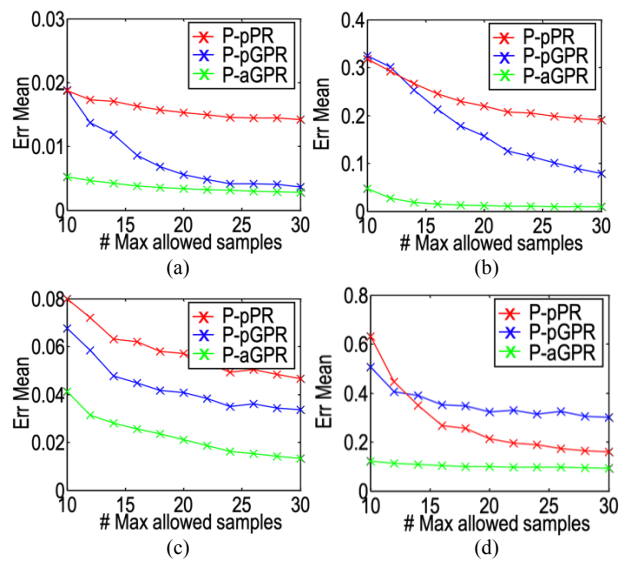

Fig. 6. The mean of absolute Euclidean distance error *Err* from 500 repeated runs for four testbench functions with different PF modeling methods, (a) ZDT1, (b) SCH, (c) SPH, (d) MAF3.

Fig. 6 shows the mean of absolute Euclidean distance error *Err* from 500 repeated runs for four testbench functions with different PF modeling methods. As shown in Fig. 6, the mean of *Err* will decrease when the maximum allowed samples $N_{max}$ increases for all three methods. Among these three PF modeling methods, the proposed P-aGPR method (i.e., the green line) achieves smallest error in four testbenches at any $N_{max}$. For instance, in SPH test case (i.e., Fig. 6(c)), the proposed P-aGPR method can achieve 0.041 absolute mean error with $N_{max} = 10$, and that of P-pPR and P-pGPR is 0.068 and 0.080 respectively. In order to achieve a similar modeling accuracy as the proposed P-aGPR method, P-pPR and P-pGPR may require $N_{max} = 30$ and $N_{max} = 20$ respectively. In this case, compared to P-pPR and P-pGPR, our proposed P-aGPR method requires 3× (i.e., 30/10 = 3) and 2× (i.e., 20/10 = 2) fewer samples respectively.

Fig. 7 shows the standard deviation (Std) of absolute Euclidean distance error *Err* from 500 repeated runs for four testbench functions with different PF modeling methods. Similar to the mean of *Err*, the proposed P-aGPR method achieves the smallest Std in all four testbenches compared to the other two methods. It implies that the proposed P-aGPR method is more stable than other two methods. The passive learning methods P-pPR and P-pGPR can be regarded as a special active learning method. Namely, trained a regression model with $N_0$ initial samples for each $L_k$, P-pPR or P-pGPR queries a new data point at random instead of using a specific criterion as P-aGPR. Therefore, they completely ignore the information already obtained in the learned models. On the contrary, the proposed P-aGPR method will wisely adjust where to query a new point based on the already obtained data set $D_k$, and thus it will be more stable.

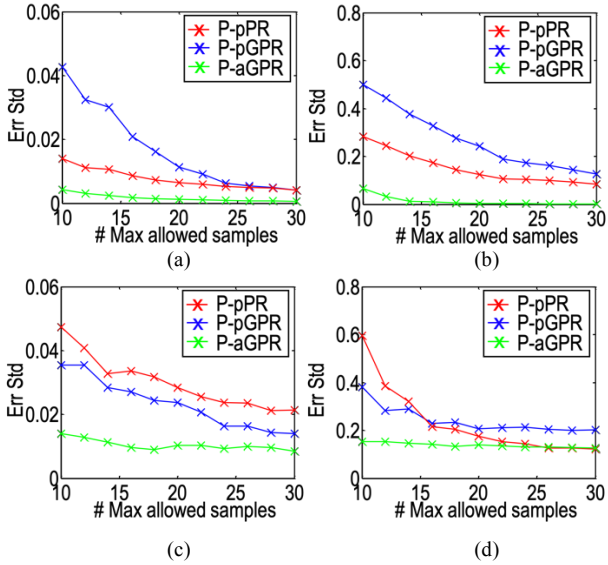

Fig. 7. The standard deviation of absolute Euclidean distance error *Err* from 500 repeated runs for four testbench functions with different PF modeling methods, (a) ZDT1, (b) SCH, (c) SPH, (d) MAF3.

Table II further summarizes the running time of three PF modeling methods with $N_{max}$ = 10 in a single run. In the bracket, we show how much more time P-aGPR will spend compared to the best case (i.e., the number denoted in bold). As shown in the first three columns, for a specific testbench, the passive methods P-pPR and P-pGPR almost spend same time. Moreover, the running time for the 3-dimensional SPH or MAF3 testbench is much larger than that of 2-dimensional ZDT1 or SCH testbench since SPH or MAF3 requires training one more regression model.

According to the last column of Table II, the running time of the proposed P-aGPR method is slightly larger than that of P-pPR or P-pGPR at the same $N_{max}$, which is unavoidable. To be more specific, the proposed P-aGPR method needs to spend time (e.g., 0.25 s, 0.26 s, 0.63 s or 0.64 s) on deciding where to query a new point, while the passive methods do not have to. In summary, at the expense of a little bit more time, the proposed P-aGPR method can achieve a far more accurate PF model compared to P-pPR and P-pGPR.

Moreover, this more time spent on querying new points can be even neglected when evaluating metrics of a design is time consuming as in real-world applications. For instance, in engineering [5]-[8], performing circuit simulation to obtain metrics of a large-scale circuit may cost days or even weeks. In this case, the time required for querying new points with maximum variance as in Eq (16) (e.g., 0.25 s, 0.26 s, 0.63 s or 0.64 s etc.) will be dominated by the time required for finding PF points (e.g., days or even weeks [5]-[8] etc.) since it needs repeatedly evaluating metrics as in Eq (11), (19) or (22).

TABLE II.
RUNNING TIME (SEC.) OF THREE DIFFERENT PF MODELING METHODS WITH $N_{MAX}$ = 10 IN A SINGLE RUN

|  | P-pPR | P-pGPR | P-aGPR (proposed) |
|---|---|---|---|
| ZDT1[22] | **0.47** | 0.50 | 0.72 (+0.25) |
| SCH[23] | 0.46 | **0.45** | 0.71 (+0.26) |
| SPH[24] | 0.93 | **0.87** | 1.50 (+0.63) |
| MAF3[25] | 1.22 | **1.17** | 1.81 (+0.64) |

Finally, we show the generated PF via the proposed P-aGPR method with $N_{max}$ = 20 for four testbench functions respectively in Fig. 8.

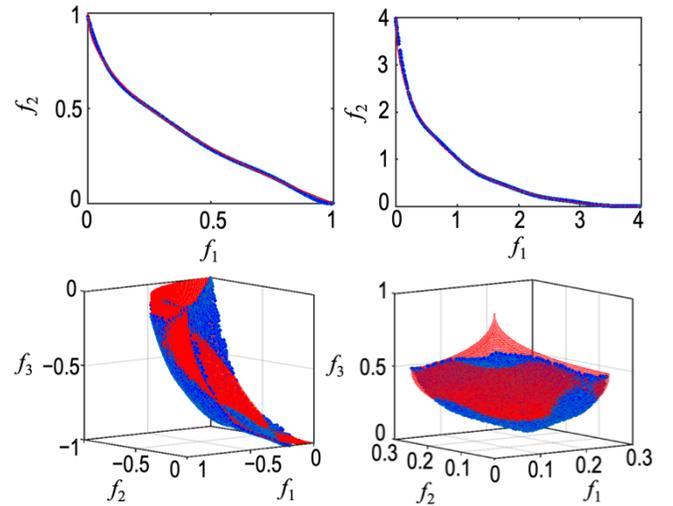

Fig. 8. Pareto Front generated via the proposed P-aGPR method with $N_{max}$ = 20. The blue dots are the PF generated with $N_{PF}$ points, and the red line or surface is the true PF. (a) ZDT1 with $N_{PF}$ = 1000, (b) SCH with $N_{PF}$ = 1000, (c) SPH with $N_{PF}$ = 8000, (d) MAF3 with $N_{PF}$ = 8000.

## VI. CONCLUSIONS

In this paper, a novel projection based active Gaussian process regression (P-aGPR) method is proposed for efficient Pareto Front (PF) modeling. P-aGPR chooses a series of projection spaces with dimensionalities ranking from low to high. Then we pose constraint for the projection of PF in each projection space with a Gaussian process regression (GPR) model. Theses GPR models are first initialized with a few PF points obtained by NBI method. Next, we query new samples with high predictive uncertainty, and consequently the GPR models will be retrained with the new queried samples. This procedure will be repeated until convergence is reached. Different from all existing methods, our proposed P-aGPR method can provide a generative model for PF and also fast examine whether a provided point locates on PF or not. The numerical results demonstrate the efficacy and stability of the proposed P-aGPR method compared to passive learning methods.

APPENDIX

We provide a proof for property (iii) $\Gamma_i = \Psi(\mathbf{g}_i(\Gamma_m))$. To prove this property, we only need to prove: (1) $\Gamma_i \subseteq \Psi(\mathbf{g}_i(\Gamma_m))$ and (2) $\Psi(\mathbf{g}_i(\Gamma_m)) \subseteq \Gamma_i$.

We begin by proving $\Gamma_i \subseteq \Psi(\mathbf{g}_i(\Gamma_m))$. Suppose that there exists $\mathbf{f}_{1:i}(\mathbf{x}) \in \Gamma_i$, i.e., $\exists \mathbf{x} \in \Omega$, and $\mathbf{f}_{1:i}(\mathbf{x}) = [f_1(\mathbf{x}) \, f_2(\mathbf{x}) \cdots f_i(\mathbf{x})]^T \in \Gamma_i$. To prove (1), we only need to prove $\mathbf{f}_{1:i}(\mathbf{x}) \in \Psi(\mathbf{g}_i(\Gamma_m))$. Now we consider $\mathbf{f}_{1:m}(\mathbf{x}) = [\mathbf{f}_{1:i}(\mathbf{x}) \, ; f_{i+1}(\mathbf{x}) \, ; f_{i+2}(\mathbf{x}) \, ; \cdots ; f_m(\mathbf{x})] = [f_1(\mathbf{x}) \, f_2(\mathbf{x}) \cdots f_m(\mathbf{x})]^T$.

The first case is that there doesn't exist any $\mathbf{f}_{1:m}(\mathbf{x}') \in \Gamma_m$ dominating over $\mathbf{f}_{1:m}(\mathbf{x})$. In this case, we conclude that $\mathbf{f}_{1:m}(\mathbf{x}) \in \Gamma_m$, and thus $\mathbf{f}_{1:i}(\mathbf{x}) = \mathbf{g}_i(\mathbf{f}_{1:m}(\mathbf{x})) \in \mathbf{g}_i(\Gamma_m)$. Furthermore, $\mathbf{f}_{1:i}(\mathbf{x})$ must contain in $\Psi(\mathbf{g}_i(\Gamma_m))$, otherwise based on the definition of $\Psi(\bullet)$ there will exist a $\mathbf{f}_{1:i}(\mathbf{x}'') \in \mathbf{g}_i(\Gamma_m) \subseteq \Theta_i$ dominating over $\mathbf{f}_{1:i}(\mathbf{x})$ and $\mathbf{f}_{1:i}(\mathbf{x}) \in \Gamma_I$ will not hold.

The second case is that there exists some $\mathbf{f}_{1:m}(\mathbf{x}') \in \Gamma_m$ dominating over $\mathbf{f}_{1:m}(\mathbf{x})$. In this case, we observe that the first $i$ entries of $\mathbf{f}_{1:m}(\mathbf{x}')$ and $\mathbf{f}_{1:m}(\mathbf{x})$ must be equal, i.e., $\mathbf{f}_{1:i}(\mathbf{x}') = \mathbf{f}_{1:i}(\mathbf{x})$, otherwise there exists a metric vector $\mathbf{f}_{1:i}(\mathbf{x}')$ dominating over $\mathbf{f}_{1:i}(\mathbf{x})$ and $\mathbf{f}_{1:i} \in \Gamma_i$ will not hold. Therefore, we obtain $\mathbf{f}_{1:m}(\mathbf{x}') \in \Gamma_m$, and thus $\mathbf{f}_{1:i}(\mathbf{x}') = \mathbf{g}_i(\mathbf{f}_{1:m}(\mathbf{x}')) \in \mathbf{g}_i(\Gamma_m)$. Furthermore, as in the first case, based on the definition of $\Psi(\bullet)$, we conclude that $\mathbf{f}_{1:i}(\mathbf{x}')$ must contain in $\Psi(\mathbf{g}_i(\Gamma_m))$. Therefore, $\mathbf{f}_{1:i}(\mathbf{x}) \in \Psi(\mathbf{g}_i(\Gamma_m))$ will also hold since we know $\mathbf{f}_{1:i}(\mathbf{x}) = \mathbf{f}_{1:i}(\mathbf{x}')$.

Up to now we have proved $\Gamma_i \subseteq \Psi(\mathbf{g}_i(\Gamma_m))$. In what follows, we will prove the other side $\Psi(\mathbf{g}_i(\Gamma_m)) \subseteq \Gamma_i$. According to what has been proved, we obtain $\Gamma_i \subseteq \Psi(\mathbf{g}_i(\Gamma_m)) \subseteq \mathbf{g}_i(\Gamma_m)$. Suppose that there exists $\mathbf{f}_{1:i}(\mathbf{x}) \in \Psi(\mathbf{g}_i(\Gamma_m))$, for some $\mathbf{x} \in \Omega$. To prove $\Psi(\mathbf{g}_i(\Gamma_m)) \subseteq \Gamma_i$, we only need to prove $\mathbf{f}_{1:i}(\mathbf{x}) \in \Gamma_i$.

Similarly, the first case is that there doesn't exist any $\mathbf{f}_{1:i}(\mathbf{x}') \in \Gamma_i$ dominating over $\mathbf{f}_{1:i}(\mathbf{x})$. In this case, we directly conclude that $\mathbf{f}_{1:i}(\mathbf{x}) \in \Gamma_i$.

The second case is that there exists some $\mathbf{f}_{1:i}(\mathbf{x}') \in \Gamma_i$ dominating over $\mathbf{f}_{1:i}(\mathbf{x})$. Since we know $\Gamma_i \subseteq \Psi(\mathbf{g}_i(\Gamma_m)) \subseteq \mathbf{g}_i(\Gamma_m)$, we can conclude $\mathbf{f}_{1:i}(\mathbf{x}') \in \mathbf{g}_i(\Gamma_m)$ dominating over $\mathbf{f}_{1:i}(\mathbf{x})$. This contradicts with the assumption $\mathbf{f}_{1:i}(\mathbf{x}) \in \Psi(\mathbf{g}_i(\Gamma_m))$. It implies that the second case will not happen.